%% file: arxiv.tex
\documentclass[nohyperref]{article}

\usepackage{microtype}
\usepackage{graphicx}
\usepackage{subfigure}
\usepackage{booktabs} 

\usepackage{hyperref}
\usepackage{balance}

\usepackage[accepted]{icml2022}

\usepackage{amsmath}
\usepackage{amssymb}
\usepackage{mathtools}
\usepackage{amsthm}

\usepackage[capitalize,noabbrev]{cleveref}

\usepackage{algorithm}
\usepackage{algorithmic}
\usepackage[utf8]{inputenc} 
\usepackage[T1]{fontenc}    
\usepackage{url}            
\usepackage{booktabs}       
\usepackage{amsfonts}       
\usepackage{nicefrac}       
\usepackage{microtype}      
\usepackage{xcolor}         
\usepackage{amsfonts}       
\usepackage{graphicx}
\usepackage{amsmath}
\usepackage{multirow}
\newcommand{\eg}{\emph{e.g., }}

\usepackage{multirow}
\usepackage{multicol}
\usepackage{threeparttable}
\usepackage{xcolor}
\usepackage{color}
\usepackage{wrapfig}
\usepackage{etoolbox} 
\usepackage{subfigure}  

\usepackage{subfigure}
\usepackage{epsfig}
\usepackage{graphicx}
\usepackage{amsmath}
\usepackage{amssymb}
\usepackage{mathrsfs}
\usepackage{bm}
\usepackage{multirow}
\usepackage{array}

\usepackage{algorithm}
\usepackage{algorithmic}

\usepackage{tcolorbox}
\usepackage[switch]{lineno}
\usepackage{booktabs}
\usepackage{xcolor}
\usepackage{colortbl}
\usepackage[switch]{lineno}
\usepackage{etoolbox} 
\usepackage{times,graphicx,amsmath,amssymb,booktabs,tabulary,multirow,overpic,xcolor}

\usepackage{algorithm}
\usepackage{algorithmic}
\usepackage{newfloat}
\usepackage{listings}
\floatstyle{ruled}
\newfloat{listing}{tb}{lst}{}
\floatname{listing}{Listing}
\newcommand*\linenomathpatch[1]{%
	\cspreto{#1}{\linenomath}%
	\cspreto{#1*}{\linenomath}%
	\cspreto{end#1}{\endlinenomath}%
	\cspreto{end#1*}{\endlinenomath}%
}

\linenomathpatch{equation}
\linenomathpatch{gather}
\linenomathpatch{multline}
\linenomathpatch{align}
\linenomathpatch{alignat}
\linenomathpatch{flalign}

\theoremstyle{plain}

\theoremstyle{definition}

\theoremstyle{remark}

\usepackage[textsize=tiny]{todonotes}

\icmltitlerunning{Unsupervised Flow-Aligned Sequence-to-Sequence Learning for Video Restoration}

\begin{document}

\twocolumn[
\icmltitle{Unsupervised Flow-Aligned \\
Sequence-to-Sequence Learning for Video Restoration}



\icmlsetsymbol{equal}{*}

\begin{icmlauthorlist}
\icmlauthor{Jing Lin}{equal,thu}
\icmlauthor{Xiaowan Hu}{equal,thu}
\icmlauthor{Yuanhao Cai}{thu}
\icmlauthor{Haoqian Wang$^{\dagger}$}{thu} \\
\icmlauthor{Youliang Yan}{huawei} 
\icmlauthor{Xueyi Zou$^{\dagger}$}{huawei}
\icmlauthor{Yulun Zhang}{ethz}
\icmlauthor{Luc Van Gool}{ethz}
\end{icmlauthorlist}

\icmlaffiliation{thu}{Shenzhen International Graduate School, Tsinghua University}
\icmlaffiliation{huawei}{Huawei Noah's Ark Lab}
\icmlaffiliation{ethz}{ETH Z\"{u}rich}


\icmlcorrespondingauthor {Xueyi Zou}{zouxueyi@huawei.com}
\icmlcorrespondingauthor {Haoqian Wang}{wanghaoqian@tsinghua.edu.cn}


\vskip 0.3in
]



\printAffiliationsAndNotice{\icmlEqualContribution} 

\begin{abstract}
How to properly model the inter-frame relation within the video sequence is an important but unsolved challenge for video restoration (VR). In this work, we propose an unsupervised flow-aligned sequence-to-sequence model (S2SVR) to address this problem. On the one hand, the sequence-to-sequence model, which has proven capable of sequence modeling in the field of natural language processing, is explored for the first time in VR. Optimized serialization modeling shows potential in capturing long-range dependencies among frames. On the other hand, we equip the sequence-to-sequence model with an unsupervised optical flow estimator to maximize its potential. The flow estimator is trained with our proposed unsupervised distillation loss, which can alleviate the data discrepancy and inaccurate degraded optical flow issues of previous flow-based methods. With reliable optical flow, we can establish accurate correspondence among multiple frames, narrowing the domain difference between 1D language and 2D misaligned frames and improving the potential of the sequence-to-sequence model. S2SVR shows superior performance in multiple VR tasks, including video deblurring, video super-resolution, and compressed video quality enhancement. \url{https://github.com/linjing7/VR-Baseline}
\end{abstract}

\vspace{-6mm}
\section{Introduction}
\label{Introduction}
Video restoration (VR) aims to reconstruct high-quality (HQ) video from its degraded low-quality (LQ) counterpart, including video deblurring~\cite{r15}, video super-resolution (SR)~\cite{r10,r25}, and compressed video enhancement~\cite{r12}. 

\begin{figure}[t!]
    \centering
    \includegraphics[width=1\linewidth]{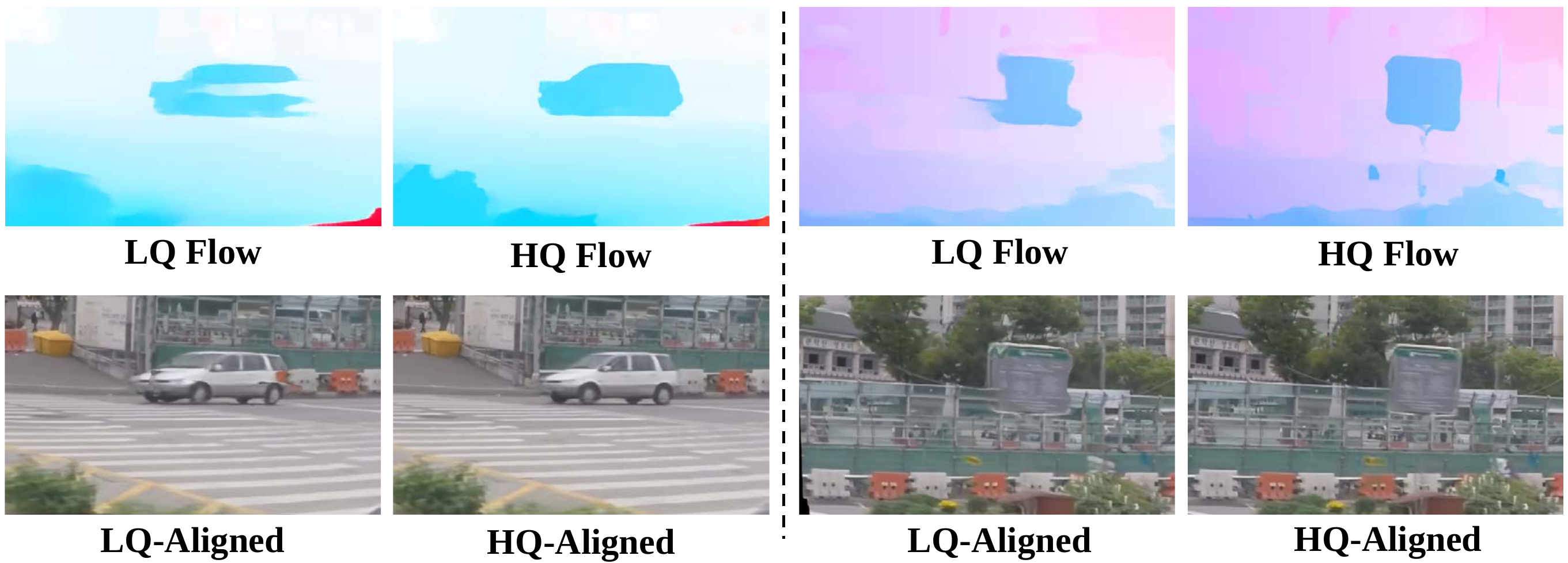}
    \vspace{-7mm} 
    \caption{Optical flow estimated from LQ and HQ videos respectively (top), and visual comparison of the aligned frames (bottom).}
    \label{fig:intro_flow}
\vspace{-6mm} 
\end{figure}

Task-driven networks often have complex structures that are elaborately designed for a specific task. These methods may be inapplicable when transferred to a new scenario or a different video restoration task~\cite{r13,deng2020spatio,cao2021video}. Therefore, it is of great significance to explore a unified and versatile framework that can be used for multiple video restoration tasks.

Early works simply extend single image restoration~\cite{r34,r35,r36,pngan} to video restoration. These image-based methods ignore inter-frame correlation, leading to limited performance. Some CNN-based methods~\cite{r11, r14, deng2020spatio} utilize information from the frames within a short temporal window. The ignorance of distant frames significantly limits the performance of these methods. Some researchers use the recurrent neural network (RNN)~\cite{r40,r24,r41,r25} to propagate the hidden state in the time domain to expand the temporal receptive field. However, as analyzed in ~\cite{jozefowicz2015empirical}, RNN suffers from both exploding and vanishing gradients. As a result, RNN is difficult to learn the long-term dependencies and can not be stacked into very deep models, limiting the representation capacity of restoration network.
The transformer-based model~\cite{cao2021video, FGST} can process a video sequence in parallel with self-attention mechanism. Nonetheless, the model complexity is quadratic to the number of tokens. For video restoration with an immense number of tokens, modeling long-range dependencies means huge computational costs and memory occupation. Thus, the problem of modeling long-term inter-frame relations with an affordable cost remains formidable. 

Based on the sequence nature of videos, our insight into this problem is to treat it as a sequence modeling task and try to solve it with the sequence-to-sequence (seq2seq) model. Seq2seq model has proven capable of sequence modeling~\cite{r8,r1,r2,r3} in the field of natural language processing (NLP), showing great potential in modeling the inter-frame relation within the video sequence. Seq2seq model
is devised to serially encode the input sequence into latent vectors and then dynamically decode a target sequence out of that representations. However, the migration of the seq2seq model is inevitably hindered by the domain discrepancy between NLP and VR. The video signal is composed of multiple misaligned 2D frames, while the seq2seq model can only handle continuous 1D input (\eg, language sequence, time series) canonically. So we need to establish accurate correspondences among multiple frames by performing a spatial alignment with optical flow estimator.

Previous flow-based~\cite{r45,r46,r47} methods perform spatial alignment with a pretrained optical flow network. \cite{r25} prove that feature alignment, $i.e.$, estimating optical flow from the LQ videos and using it to warp the hidden state, can yield a better restoration result than image alignment. However, these flow-based methods may be suboptimal and suffer from the following issues: Firstly, the data discrepancy between synthetic flow dataset and real-world video affects the performance of the pretrained optical flow module in VR. Secondly, the optical flow estimated from the LQ input video (LQ flow) may be unreliable since the video degradation may seriously distort video contents and break pixel-wise correspondences between frames~\cite{zheng2021adaptive}. As shown in Fig.~\ref{fig:intro_flow}, the LQ flows lose some motion details, and the frames aligned by the LQ flows (LQ-aligned frames) contain blurry edges. In contrast, the HQ flow is more detailed,  and the HQ-aligned frames contain sharper semantics. Besides, for feature alignment, the motion information estimated from the LQ video may be inconsistent with that of the hidden state, which is expected to be spatially aligned with the HQ video. So some artifacts will be brought when the LQ flow is used for feature alignment.

We attempt to address the data discrepancy and inaccurate LQ flow issues with unsupervised distillation optical flow loss. To be specific, we train an optical flow estimator on the VR dataset with unsupervised loss. The data discrepancy naturally disappear since the training and testing dataset both come from the real-world VR dataset. Furthermore, a novel data distillation loss is designed to generate more accurate LQ flows, in which the optical flows estimated from the HQ video serve as the pseudo-labels of the LQ flows. This loss encourages the LQ flows to imitate the HQ flows, which are more accurate and  spatial consistent with the motion information of the hidden state.

Therefore, the unsupervised flow-aligned sequence-to-sequence model is proposed for video restoration tasks (S2SVR). We migrate and improve the seq2seq model from NLP to VR task, and maximize the potential of the seq2seq model with an unsupervised optical flow estimator. In a nutshell, our contributions can be summarized as follows:

\vspace{-1mm}
\begin{itemize}
    \vspace{-3.0mm}
    \item This is the first VR work to explore the sequence-to-sequence model, which comes from NLP and is intrinsically suitable for video sequence modeling.
    \vspace{-2.0mm}
    \item The proposed unsupervised distillation optical flow loss alleviates the data discrepancy and inaccurate LQ flow issues of previous flow-based methods, narrowing the domain difference between NLP and VR.
    \vspace{-2.0mm}
    \item Extensive experiments show that our method achieves state-of-the-art performance in three typical video restoration tasks, including video deblurring, video super-resolution, and compressed video enhancement. 
\end{itemize}
\vspace{-5mm}

\section{Related Work}
\vspace{-1mm}
\subsection{Video Restoration} 
\vspace{-1mm}
Early work~\cite{r33,r35,r34} adopt an image restoration model for video restoration and do not take advantage of information in the neighbouring frames. The ignorance of the inter-frame correlation severely limits the restoration result. Some CNN-based methods~\cite{deng2020spatio,tian2018tdan} employ deformable convolution to perform feature-level alignment. The RNN-based methods design the recurrent structure and attempt to model the long-term dependencies by propagating the hidden state~\cite{r40,r24,r41}. 
~\cite{r25} prove that the combination of bidirectional propagation and optical flow estimation can achieve ideal results.~\cite{deng2021multi} propose a recurrent model with separable-patch architecture and multi-scale integration scheme for fast and accurate video deblurring. However, the RNN-based methods inevidently suffer from the vanishing gradient problem and have difficulty in capturing the long-range temporal dependencies. Recently, the emerging Transformer model has been applied in image and video restoration tasks~\cite{MST,VRT,CST,cao2021video,MST++}. Nonetheless, the token-based self-attention module has enormous computational and memory cost in restoring long video sequence. Thus, the problem of effectively modeling long-range temporal dependencies within the video sequence remains formidable.
\begin{figure*}[htp!]
    \centering
    \includegraphics[width=1\linewidth]{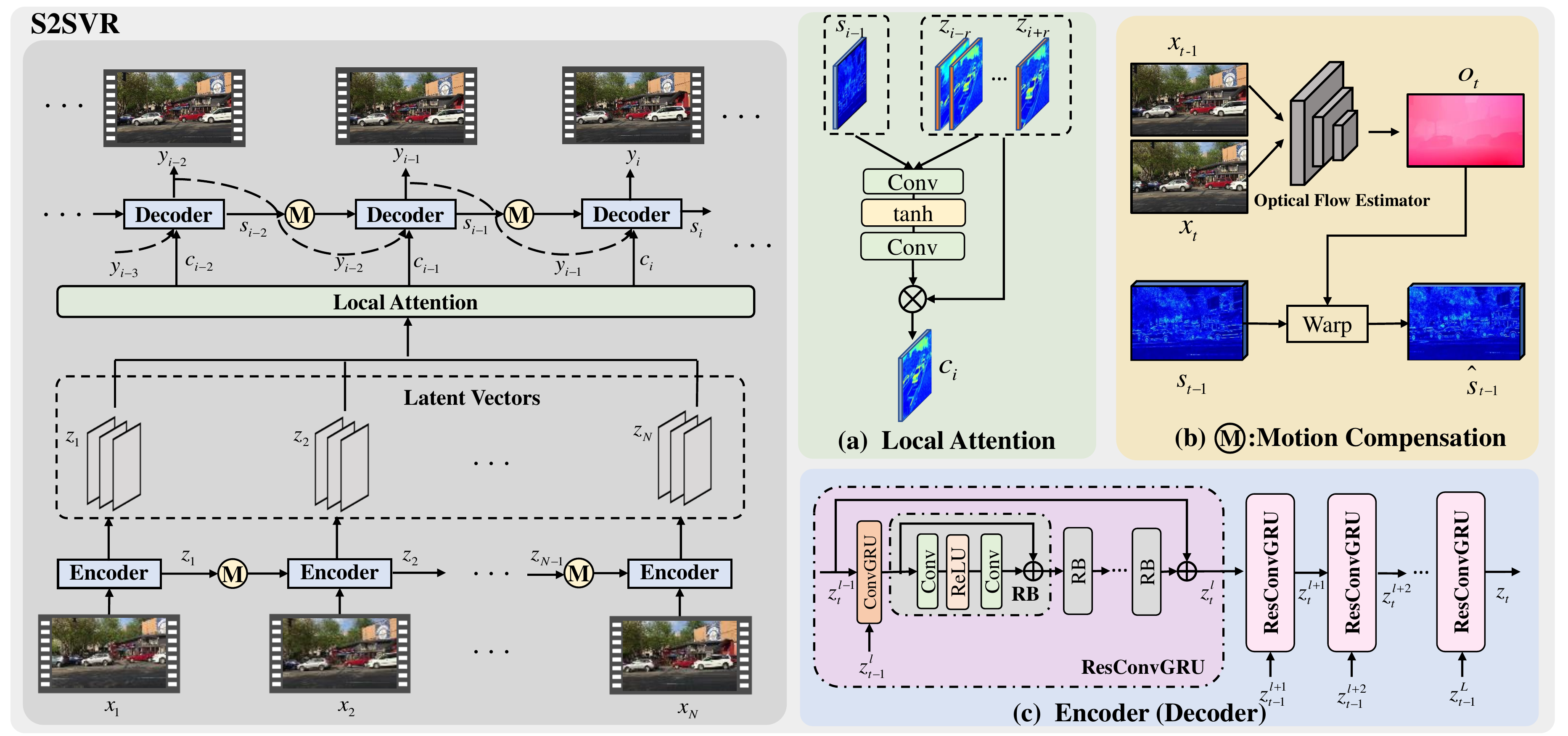}
    \vspace{-8mm} 
    \caption{The architecture of the proposed unsupervised flow-aligned seq2seq model (S2SVR). The modules with different background colors on the right show the internal details of (a) local attention, (b) motion compensation  and (c) Encoder (Decoder).}
    \label{pipeline}
\vspace{-3.5mm} 
\end{figure*}
\vspace{-3mm}
\subsection{Sequence-to-Sequence Learning}
\vspace{-2mm}
Seq2seq model is first proposed by \cite{r8} for the machine translation task in which a long short-term memory (LSTM) encodes the input sequence into a latent representation and then another LSTM decodes the target sequence out of that representation. The model is intrinsically suitable for long-range coding tasks. Various variants of the seq2seq model have been applied to many sequence modeling tasks, such as speech recognition \cite{r42}, time series analysis \cite{r31, r30}, and text summarization~\cite{shi2021neural}. Due to the fundamental difference between video and language, the potential of this serialized encoding-decoding structure in assisting continuous-frame VR is unexplored.

\vspace{-3.5mm}
\subsection{Optical Flow Estimation}
\vspace{-1.5mm}
With the development of deep learning, some optical flow estimation networks~\cite{r46,r47} trained on synthetic datasets have achieved better results than non-learning methods~\cite{r44,r45}. The domain difference between synthetic optical flow and real-world optical flow datasets leads to limited model performance. \cite{r49} suggest using an unsupervised optical flow estimator to circumvent the need for labels. \cite{wang2018occlusion} improve the performance of unsupervised optical methods by proposing a new warping module to facilitate large motion learning and model occlusion explicitly.~\cite{r16} train a task-oriented flow module jointly with the video enhancement module in the supervision of $\mathcal{L}_1$ loss. But the jointly-trained flow module becomes unsuitable when cooperating with other video processing modules. Besides, they have not solved the problem that it's difficult to estimate accurate motion information from the severely degraded input frames. Based on the LQ-HQ paired characteristics of VR tasks, we propose a data distillation loss to improve the quality of the LQ flows.  

\vspace{-1mm}
\section{Method}
\vspace{-2mm}
In this section, we present our S2SVR model. We first introduce the overall framework of the seq2seq model. Then, we explain the unsupervised distillation optical flow method, which narrow the domain discrepancy between NLP and VR and improve the potential of the seq2seq model in VR.
\vspace{-2mm}
\subsection{Sequence-to-Sequence Learning}
\vspace{-1mm}
To promise that the scalable seq2seq architectures and their efficient implementations can be preserved, S2SVR follows the seq2seq framework from NLP as closely as possible. As shown in Fig.~\ref{pipeline}, S2SVR is composed of four components: encoder, decoder, local attention, and optical flow estimator. 

For notation, we use capital letters to represent sequences,(\eg $X$,$Y$), lower case to denote individual frames in a sequence, (\eg $x_1$,$x_2$). Let $X=\{x_1,x_2,\ldots,x_N\}$ represent the input low-quality video sequence and $Y=\{y_1,y_2,\ldots,y_N\}$ be the corresponding high-quality video sequence, where $N$ is the length of the sequence. The goal of our S2SVR is to estimate the conditional probability of the target sequence respective to the input sequence $P(Y|X)$.

\textbf{Encoder.}~Firstly, the encoder read sequentially each $x_i \in X$ and transforms the source sequence into a list of latent vectors $Z=\{z_1,z_2,\ldots,z_N\}$:
\begin{equation}
\label{eq:encoder}
z_i = F_e(z_{i-1},x_i),
\end{equation}
where $z_i$ denotes the latent vector at time step $i$, and $F_e$ denotes the function of the encoder, which in our implementation is a residual stacked ConvGRU (ResConvGRU). The ResConvGRU will be introduced in the next subsection.

\textbf{Decoder.} Next, the decoder sequentially produces the output video based on the encoded vectors. Specifically, using the chain rule, the conditional probability $P(Y|X)$  can be decomposed as:
\begin{equation}
\begin{aligned}
P(Y|X) &= P(y_1, y_2, \ldots, y_{N}|z_1,z_2,\ldots,z_N) \\
&=\prod_{t=1}^{N} P(y_t|y_1, \ldots, y_{t-1};z_1,\ldots,z_N).
\end{aligned}
\end{equation}
We serially generate the subsequent output based on the source sequence encoding and the decoded sequence so far:
\begin{equation}
\label{eq:decoder}
y_t = F_d(y_1, y_2, \ldots, y_{t-1};z_1,z_2,\ldots,z_N).
\end{equation}
$F_d$ represents the decoder, which is composed of a ResConvGRU and a feed-forward network. ResConvGRU generates a hidden state $s_i$, and then $s_i$ passes through the feed-forward network to produce the output frame:
\begin{equation}
\begin{aligned}
\label{eq:decoding}
s_i &= F_r(s_{i-1}, y_{i-1},c_i), \\
y_i &= F_f(s_i),
\end{aligned}
\end{equation}
where $F_r$ is the ResConvGRU and $F_f$ denotes the feed-forward network. $s_i,y_i$ refer to the hidden state of ResConvGRU and output frame at $i^{th}$ time step, respectively.  And $c_i$ is a context vector generated by the local attention module based on the latent vectors $Z=\{z_1,z_2,\ldots,z_N\}$.

\textbf{Local Attention.} As shown in Fig.{~\ref{pipeline}(a)}, the attention module generates a context vector $c_i$ for each time step, allowing the decoder to extract information from different parts of the input sequence. Specifically, we represent the context vector $c_i$ as a weighted sum of a subset of the latent vectors:
\begin{equation}
\label{eq:context vector}
c_i = \sum_{j={i-r}}^{i+r}\alpha_{ij}z_j,
\end{equation}
where $r$ is the the subset radius and the weight $\alpha_{ij}$ is:
\begin{equation}
\label{eq:vid_res}
    \alpha_{ij} = \frac{exp(e_{ij})}{\sum_{k=i-r}^{i+r}exp(e_{ik})}.  \\
\end{equation}
$e_{ij} = F_a(s_{i-1},z_j)$ is an attention model scoring the correspondence between the $i^{th}$ input and the $j^{th}$ output based on $s_{i-1}$ and $z_j$. Similar to~\cite{r35}, a two-layer feed-forward network is adopted as the attention model:
\begin{equation}
\label{eq:local_attention}
e_{ij}= \mathbf{V}_a \cdot tanh(\mathbf{W}_a[s_{i-1},z_j]),
\end{equation}
where $\mathbf{V}_a$ and $\mathbf{W}_a$ denote the first and second convolution layers of the feed-forward network, respectively. And $[\cdot ,\cdot]$ refers to concatenation along the channel dimension. 

\textbf{Motion Compensation.} To improve the performance of the seq2seq model in VR, we need to establish accurate spatial correspondences among multiple frames. Similar to previous methods \cite{r40,r24,r41,r25}, we adopt an optical flow estimator for motion compensation. Specifically, as shown in Fig.{~\ref{pipeline}(b)}, we employ a flow estimator to predict the motion between two consecutive frames. Then we warp the hidden state of ResConvGRU at last time step $s_{t-1}$, making it spatially aligned with the input at the current step:
\begin{equation}
\begin{aligned}
\label{eq:motion_compensation}
o_t &= F_o(x_t,x_{t-1}), \\
\hat{s}_{t-1} &= F_w(s_{t-1}, o_t),
\end{aligned}
\end{equation}
where $F_o$ and $F_w$ respectively refer to the optical flow estimator and  spatial warping module. $o_t$ is the optical flow field between the adjacent input frames $x_t$ and $x_{t-1}$.
\begin{figure}[t!]
    \centering
    \includegraphics[width=0.95\linewidth]{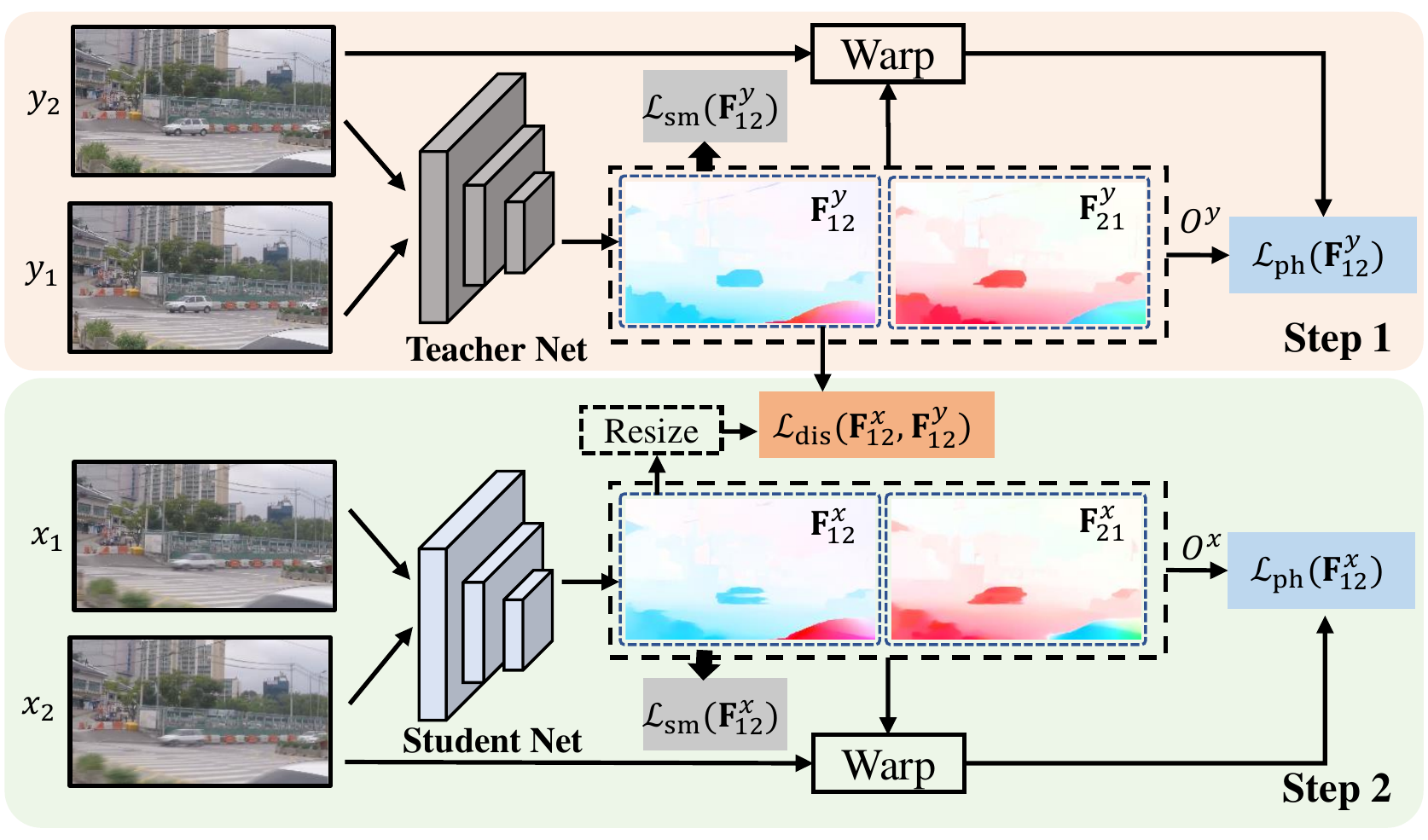}
    \vspace{-2.5mm} 
    \caption{Illustration of the unsupervised optical flow method.}
    \label{fig:method_flow}
\vspace{-3mm} 
\end{figure}
\vspace{-3mm}
\subsection{Residual Stacked ConvGRU}
\vspace{-1mm}
We use a deep-stacked ConvGRU for both the encoder and the decoder. Considering the video characteristics, as shown in Fig.~\ref{pipeline}(c), we make two modifications to the original ConvGRU. Firstly, to improve the image processing ability, several residual blocks are concatenated after the ConvGRU. Besides, motivated by the idea of modeling the difference between an intermediate layer's output and the target, we introduce residual among the layers in a stack. We define the ConvGRU and residual blocks as $F_g(\cdot)$ and $F_b(\cdot)$:
\begin{equation}
\label{eq:residual rnn}
z_t^{l} = z_t^{l-1}+F_b(F_g(z_{t-1}^l,z_t^{l-1})),
\end{equation}
where $z_{t}^l$ denote the hidden state of $l^{th}$ ConvGRU at time step $t$.
In this way, the vanishing gradient problem  can be addressed, allowing us to model the long-term temporal dependencies. More details are provided in the appendix. 
\vspace{-3mm}
\subsection{Unsupervised Optical Flow Estimator}
\vspace{-1mm} 
As analyzed in Sec.~\ref{Introduction}, previous flow-based motion compensation methods suffer from the data discrepancy between synthesized and real-world datasets, as well as inaccurate LQ flows. To solve these problems, we propose an unsupervised scheme equipped with a novel distillation loss to train the flow estimator on the VR dataset as shown in Fig.~\ref{fig:method_flow}.

Let $X$ denotes a LQ input video, and $Y$ is the corresponding HQ video. Our goal is to train a flow network $F_o$ that can estimate accurate motion information from the LQ videos (HQ videos are unavailable during inference) by predicting the optical flow $\mathbf{F}_{12}^x$ for two consecutive LQ frames $\{x_1,x_2\}$:
\begin{equation}
\label{eq:flow esimator}
\mathbf{F}_{12}^x = F_o(x_1,x_2).
\end{equation}

\begin{figure*}[t!]
\centering
\includegraphics[width=0.96\linewidth]{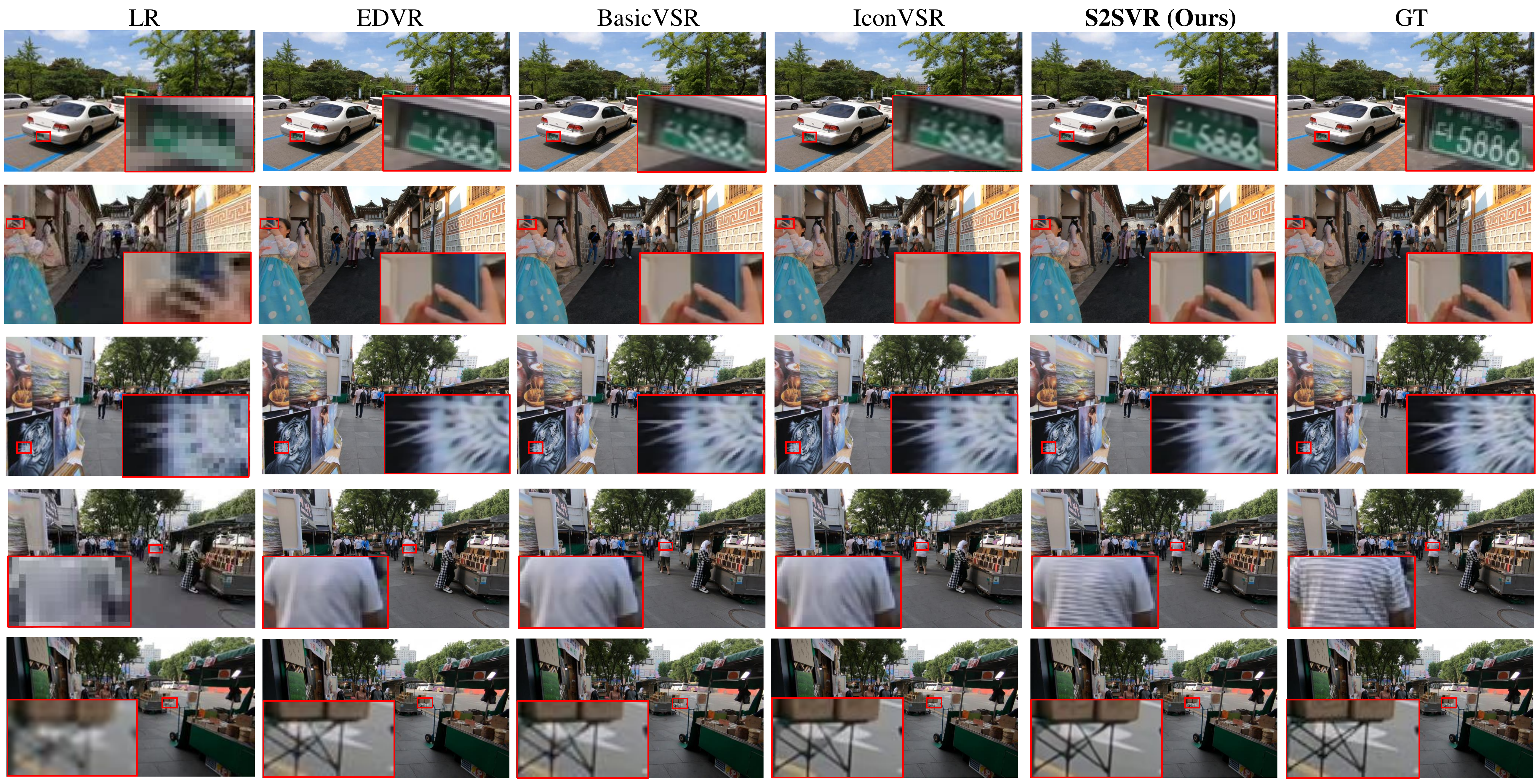}
\vspace{-4.5mm}
\caption{Visual comparison of video 4$\times$ SR results on the REDS4~\cite{r58} dataset. Please zoom in for a better comparison.}
\label{SR_vis}
\vspace{-3mm}
\end{figure*}
\vspace{-2mm}
The unsupervised scheme is summarized in Algorithm~\ref{alg:1} to better understand the proposed unsupervised optical flow estimation method. Firstly, we train a teacher flow estimation network parameterized by $\theta_t$ on the HQ videos with photometric loss and smooth loss. After convergence, we use the pretraied teacher estimator to generate pseudo-labels and train a student flow network parameterized by $\theta_o$ on the LQ video. In the following, we explain the proposed unsupervised optical flow training scheme step by step.

\textbf{Step 1.} We train an optical flow estimator $F_{t}$ with photometric loss and smooth loss on the HQ video $Y$. This optical flow estimator $F_{t}$ will be frozen and serves as a teacher network in the next step. The photometric loss~\cite{photometric_loss} is based on the assumption that the same object in two consecutive frames must have similar intensities:
\begin{equation}
\label{eq:pho_loss}
\mathcal{L}_{\text{ph}}(\mathbf{F}_{12}^y) = \sum_p \rho(y_1(p),y_2(p+\mathbf{F}_{12}^y(p)))\cdot O^y(p),
\end{equation}
where $p$ is the coordinate and $O^y$ is the occlusion mask to discard the loss on the occurred region generated by the bidirectional checking~\cite{wang2018occlusion}, $\rho(\cdot)$ is the  $\ell_1$ loss, and $\mathbf{F}_{12}^y$ is the optical flow field for two consecutive frames in the HQ videos $Y$:
\begin{equation}
\label{eq:teacher_flow_estimator_2}
\mathbf{F}_{12}^y = F_t(y_1,y_2).
\end{equation}

Further, we adopt a one-order smooth loss~\cite{smooth_loss} to encourage collinearity of neighboring flows:
\begin{equation}
\label{eq:smooth_loss}
\mathcal{L}_{\text{sm}}(\mathbf{F}_{12}^y) = \sum_{d \in x, y} \sum_{p} |\partial_{d} \mathbf{F}_{12}^y(p)| e^{-\left|\partial_{d} y_1(p)\right|}
\end{equation}
And then we formulate the loss used in the first step as:
\begin{equation}
\label{eq:loss_step1}
\mathcal{L} = \omega_{\text{ph}}\cdot\mathcal{L}_{\text{ph}}(\mathbf{F}_{12}^y)+\omega_{\text{sm}}\cdot \mathcal{L}_{\text{sm}}(\mathbf{F}_{12}^y).
\end{equation}
We respectively set the weights $\omega_{\text{ph}}$ and $\omega_{\text{sm}}$ to 0.15 and 50.

\input{algorithm}

\textbf{Step 2.} Now we have trained a teacher optical flow estimator $F_t$ which can predict the accurate optical flow $\mathbf{F}_{12}^y$ for two consecutive HQ frames $\{y_1,y_2\}\in Y$:
\begin{equation}
\begin{aligned}
\label{eq:teacher_flow_estimator}
\mathbf{F}_{12}^y &= F_t(y_1,y_2). \\
\end{aligned}
\end{equation}
Based on the assumption that the HQ flow is more accurate for motion compensation, we use $\mathbf{F}_{12}^y$ as the pseudo-labels of the LQ flows $F_{12}^x$ and and propose the distillation loss:
\begin{equation}
\label{eq:data-distillation loss}
\mathcal{L}_\text{dis}(\mathbf{F}_{12}^x,\mathbf{F}_{12}^y) = \sum_p |{\mathbf{F}}_{12}^y(p)-{F_u(\mathbf{F}}_{12}^x)(p)|,
\end{equation}

where $F_u$ is a upsample operation to ensure that $\mathbf{F}_{12}^x$ has the same size as $\mathbf{F}_{12}^y$ in video super-resolution task.
Along with the photometric loss and smoothness regularization, we train the student flow estimator $F_o$ on the LQ dataset:
\begin{equation}
\label{eq:total_loss}
\mathcal{L} = \omega_{\text{ph}}\mathcal{L}_{\text{ph}}(\mathbf{F}_{12}^x)+\omega_{\text{sm}} \mathcal{L}_{\text{sm}}(\mathbf{F}_{12}^x)+\omega_{\text{dis}}\mathcal{L}_\text{dis}(\mathbf{F}_{12}^x,\mathbf{F}_{12}^y).
\end{equation}
We set the weights to \{$\omega_{\text{ph}}=0.15,\omega_{\text{sm}}=50,\omega_{\text{dis}}=0.1$\}.
The student network will be later used as our optical flow estimator for motion compensation as in Eq.~\eqref{eq:motion_compensation}. In implementation, we adopt a lightweight flow model pwclite~\cite{r55} as our optical flow network.

\vspace{-3mm}
\section{Experiments}
\vspace{-2mm}
\subsection{Implementation Details}
\vspace{-1mm}
\noindent\textbf{Datasets.}  For video SR, the benchmark datasets consist of REDS4~\cite{r58} and Vimeo-90K-T \cite{r19}. For video deblurring, we use the GOPRO dataset~\cite{r74}, where 22 videos are used for training and 11 videos for testing. For compressed video enhancement, our models are trained with the MFQEv2 dataset \cite{r12} including 108 lossless videos. We adopt the dataset from ITU-T \cite{r57} containing 18 videos for evaluation. We compress videos by HEVC reference software HM16.5 under Low Delay P (LDP) configuration~\cite{r12,deng2020spatio}. Evaluation metrics include PSNR and SSIM \cite{r60}.

\noindent\textbf{Settings.} Models are trained with nature videos and their degraded counterparts. During unsupervised optical flow training, the learning rate is set to $1\times10^{-4}$. And during restoration training, the initial learning rate of the flow estimator and the other modules are set to $5\times10^{-5}$ and $2\times10^{-4}$, respectively. We use PyTorch to implement our models and train them on 8 Tesla V100 GPUs. More details are provided in the supplementary material due to space limitation.

\input{tables/vsr}

\input{tables/video_deblur}
\input{tables/vqe}

\begin{figure*}[t!]
\vspace{-2mm}
\centering
\includegraphics[width=0.97\linewidth]{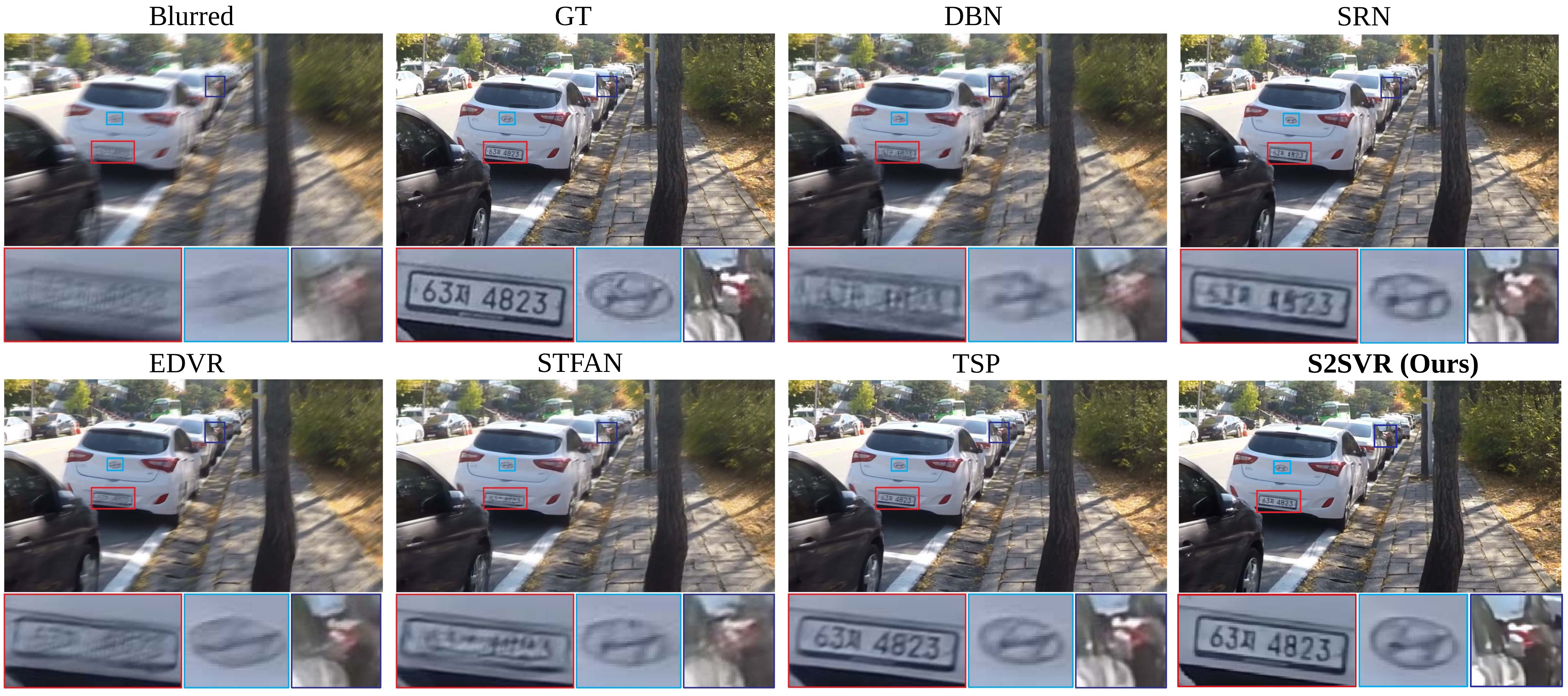}
\vspace{-5mm}
\caption{Visual comparison of video deblurring results on the GOPRO~\cite{r74} dataset. Please zoom in for a better comparison.}
\label{deblur_vis}
\vspace{-3mm}
\end{figure*}

\vspace{2mm}
\subsection{Video Super-Resolution}
\vspace{-1mm}
\noindent\textbf{Quantitative Comparison.} We compare our method with previous methods: TOFlow \cite{r19}, DUF \cite{r75}, RBPN \cite{r76}, EDVR-M \cite{r11}, EDVR \cite{r11}, PFNL \cite{r77}, MuCAN \cite{r78}, BasicVSR \cite{r25}, IconVSR \cite{r25}, and VSR-Transformer~\cite{cao2021video}. As shown in Tab.~\ref{vsr_quan}, it is clear that our method outperforms all other models by a large margin on the REDS4 dataset. Specifically, our S2SVR model achieves 0.29dB gain over the suboptimal model and 0.77dB over the VSR-Transformer model in PSNR. For Vimeo-90K-T, our performance is slightly lower than VSR-Transformer, but S2SVR only requires 41\% parameters compared with the latter. It shows that we only need half the parameters to obtain comparable performance to transformer-based models. Note that Vimeo-90K-T contains sequences with seven frames. So it also indicates that our method performs better in restoring long sequences. Serialized modeling of seq2seq models and accurate optical flow estimation facilitates the capture of long-range inter-frame dependencies.

\noindent\textbf{Visual Comparison.} From the comparison with other methods in Fig.~\ref{SR_vis}, our S2SVR network has shown great advantages in the restoration of textures and structural details, such as license plate numbers, pane lines, and hairs. Our results are more reliable and detailed, while the other methods suffer from excessive smoothing and content distortion.

\vspace{-3mm}
\subsection{Video Deblurring}
\vspace{-2mm}
\noindent\textbf{Quantitative Comparison.}
We compare our method against state-of-the-art algorithms, including Tao \textit{et al.} \cite{r67}, Su \textit{et al.} \cite{r68}, Kim \textit{et al.} \cite{r70}, Nah \textit{et al.} \cite{r71}, EDVR \cite{r11}, STFAN \cite{r72}, TSP \cite{r14}, and UHDVD~\cite{deng2021multi}. The Tab.~\ref{tab:GOPRO_result} shows the quantitative results on the GOPRO dataset~\cite{r74}. Our proposed method performs favorably against other methods and has an absolute advantage on PSNR in video deblurring. Specifically, the S2SVR model achieves a performance gain of 0.14dB on the dataset with a lightweight structure. We also report the size of the open-source model in Tab.~\ref{tab:GOPRO_result}. As the largest model, EDVR's parameter is up to 23M, but its performance is unsatisfactory. Our S2SVR network contains 8.44M parameters. Compared with the TSP~\cite{r14}, our model achieves a higher PSNR performance with only one-half of its size. 

\vspace{-1mm}
\noindent\textbf{Visual Comparison.} From the comparison results in Fig.~\ref{deblur_vis}, it can be seen that our method can restore the original structure as much as possible from the severely degraded scene. Digital restoration of blurred scenes is difficult. It can be seen that no other method except ours can guarantee the semantics while still retaining the satisfying visual results.

\begin{figure}
    \includegraphics[width=0.99\linewidth]{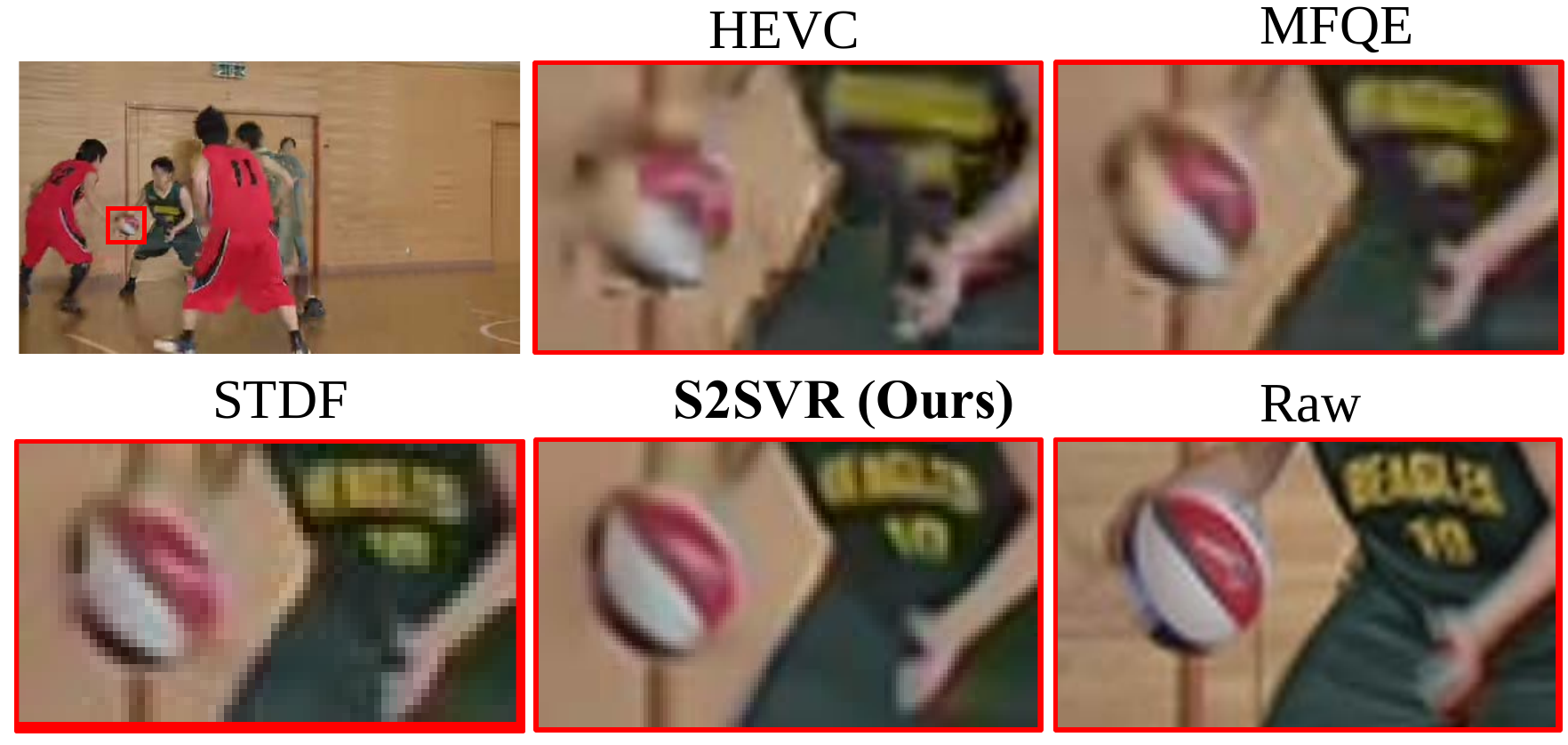}
    \vspace{-4.5mm}
    \caption{Visual comparison on Video BasketballPass at QP = 37.}
\label{compress_vis}
\vspace{-3mm}
\end{figure}

\begin{figure}
\vspace{-2mm} 
    \includegraphics[width=0.99\linewidth]{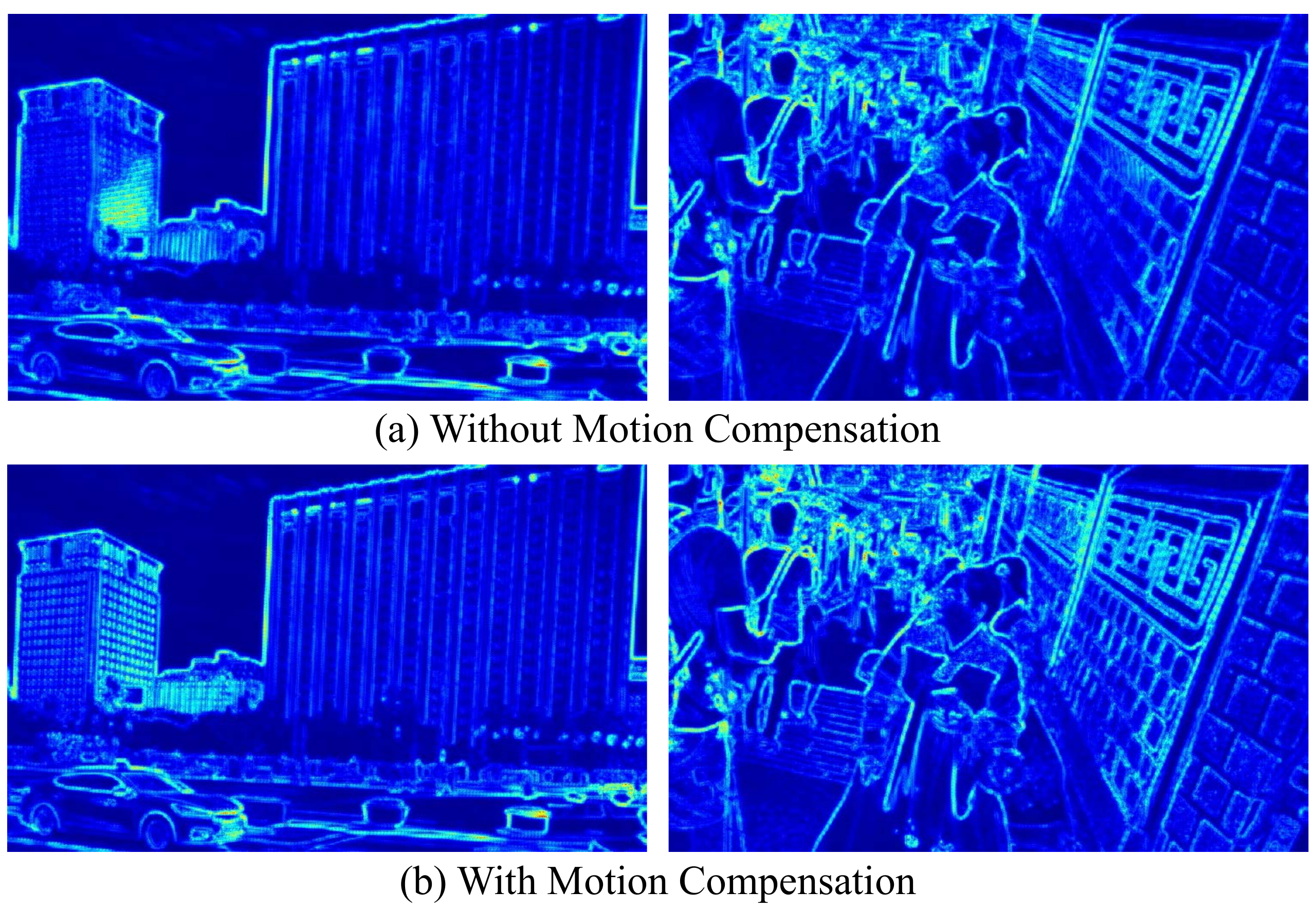}
    \vspace{-4.5mm}
    \caption{Visualization with and without motion compensation.}
\label{motion_compensation}
\vspace{-7mm}
\end{figure}

\vspace{-3mm}
\subsection{Compressed Video Enhancement}
\vspace{-2mm}
\noindent\textbf{Quantitative Comparison.}
We evaluate the performance of compressed video enhancement by $\Delta$PSNR and $\Delta$SSIM, which measure the PSNR and SSIM improvement after the enhancement. We compare S2SVR with AR-CNN \cite{dong2015compression}, DnCNN \cite{Zhang2017Beyond}, DS-CNN \cite{yang2018enhancing}, MFQE 1.0 \cite{r13}, MFQE 2.0 \cite{r12}, and STDF-R3L \cite{deng2020spatio}. As shown in Tab.~\ref{tab:vqe}, S2SVR almost outperforms all compared methods in $\Delta$PSNR and $\Delta$SSIM by a large margin. On the $RaceHorses$ dataset with the 832 $\times$480 input size, our method can outperform the latest method by at least 0.304dB, which shows the superiority of our method in processing long video sequences with large motion. 
\vspace{-1mm}

\noindent\textbf{Visual Comparison.} As shown in Fig.~\ref{compress_vis}, some structural distortions and color deviations make the results of previous methods unconvincing. Our S2SVR network guarantees the basic structural texture and semantic content. Due to space limitations, we put more comparisons in the supplementary.

\vspace{-3mm}
\section{Ablation Study}
\vspace{-1mm}
\noindent\textbf{Unsupervised Optical Flow Estimator.}
To demonstrate the effectiveness of the unsupervised training scheme, we retrain our optical flow network pwclite in a supervised manner with the optical flow dataset FlyingChairs~\cite{r80}. We also adopt pre-trained RAFT~\cite{r47}, the SOTA supervised optical flow network, as our optical flow estimator. As shown in Tab.~\ref{tab:unsupervised_optical}, the pwclite trained with our unsupervised distillation loss can outperform the supervised counterpart by 0.17 dB. It indicates that the flow estimator trained in our unsupervised scheme fits the VR tasks well. Notably, it achieves a better result than the state-of-the-art supervised method RAFT by 0.08dB with a lower cost. These results show the effectiveness of our unsupervised optical flow method.

\input{tables/ablation}
\input{tables/Length}

\vspace{-1mm}
\noindent\textbf{Motion Compensation Visualization.}
We visualize the feature saliency maps with (W/I) and without (W/O) motion compensation in Fig.~\ref{motion_compensation}. Obviously, the video frame will lose lots of motion details and texture edges without motion compensation. It is caused by the misalignment among multiple frames, which limits the potential of the seq2seq  model in VR. In contrast, with motion compensation, the feature map is much sharper and preserves more movement details, which benefits from our accurate optical flow estimation. Motion compensation narrows the domain difference between NLP and VR, facilitating information aggregation.

\vspace{-1mm}
\noindent\textbf{Long Sequence Reconstruction.} 
To validate the effectiveness of our S2SVR in capturing long-range temporal dependencies, we separate a video in REDS4 dataset into 4 segments with different lengths, including $5,15,30,$ and $50$ frames, respectively. And we use S2SVR, EDVR, and EDVR-M to restore these sequences independently. Our method performs the best among the three methods in Tab.~\ref{tab:length}. And the longer the sequence is, the more superior our S2SVR shows. It suggests that our method has an excellent performance in modeling long-range dependencies.

\vspace{-3mm}
\section{Conclusions}
\vspace{-1mm}
In this paper, we propose an unsupervised flow-aligned seq2seq model for multiple video restoration tasks. Our work aims at solving the challenges of properly modeling the inter-frame relation within the video sequence. The sequence-to-sequence learning is explored for the first time in VR to capture long-term temporal dependencies at a low cost. What's more, we design an unsupervised optical method equipped with a novel distillation loss to improve the performance of the seq2seq model in VR. Extensive experiments show that the proposed method achieves comparable performance in video deblurring, video super-resolution, and compressed video quality enhancement tasks with moderate model size, especially in long sequence VR.

\noindent\textbf{Acknowledgements:} 
This work is partially supported by the NSFC fund (61831014), the Shenzhen Science and Technology Project under Grant (CJGJZD20200617102601004, JSGG20210802153150005).

\bibliography{icml22_ref}
\bibliographystyle{icml2022}

\end{document}

%% file: algorithm.tex
\begin{algorithm}[t]
\caption{Unsupervised Distillation Optical Flow Loss}
\label{alg:1}
\begin{algorithmic}
\STATE \textbf{Inputs:} teacher and student net parameterized by $\theta_t,\theta_o$, \\
cost function parameters: loss weights \{$\omega_{\text{ph}},\omega_{\text{sm}},\omega_{\text{dis}}$\}, \\
optimization parameters: number of iterations $T$
\STATE \textbf{Output:} pretrained student network \\
// Step1: train the teacher network \\
\FOR{$j=0$ to $T$}
\STATE compute photometric loss $\mathcal{L}_{\text{ph}}$ (using Eq.~\eqref{eq:pho_loss})
\STATE compute smooth loss $\mathcal{L}_{\text{sm}}$ (using Eq.~\eqref{eq:smooth_loss})
\STATE $\mathcal{L}_{\text{tot}}=\omega_{\text{ph}}\cdot\mathcal{L}_{\text{ph}}+\omega_{\text{sm}}\cdot \mathcal{L}_{\text{sm}}$, ~~~$\nabla L(\theta_t) = \frac{\partial L_{tot}}{\partial \theta_t}$,
\STATE $\theta_t = \theta_t - \alpha \nabla L(\theta_t)$
\ENDFOR\\
// Step2: train the student network \\
\FOR{$j=0$ to $T$}
\STATE compute photometric loss $\mathcal{L}_{\text{ph}}$ (using Eq.~\eqref{eq:pho_loss})
\STATE compute smooth loss $\mathcal{L}_{\text{sm}}$ (using Eq.~\eqref{eq:smooth_loss})
\STATE compute data distillation loss $\mathcal{L}_{\text{dis}}$ (using Eq.~\eqref{eq:data-distillation loss})
\STATE $\mathcal{L}_{\text{tot}} = \omega_{\text{ph}}\mathcal{L}_{\text{ph}}+\omega_{\text{sm}} \mathcal{L}_{\text{sm}}+\omega_{\text{dis}}\mathcal{L}_\text{dis}$, ~$\nabla L(\theta_o) = \frac{\partial \mathcal{L}_{tot}}{\partial \theta_o}$,
\STATE $\theta_o = \theta_o - \alpha \nabla L(\theta_o)$
\ENDFOR
\STATE \textbf{Return} student network parameters~~$\theta_o$
\end{algorithmic}
\end{algorithm}


%% file: tables/vsr.tex
\vspace{-0.4cm}
\begin{table}[!t]
	\vspace{-0.1cm}
	\begin{center}\scalebox{0.95}{
			\tabcolsep=0.1cm
			\begin{tabular}{l|r|c|c}
				\hline                                                           
				                             Methods   & Params & REDS4            & Vimeo-90K-T  \\ \hline  
				Bicubic                         & -                  & 26.14 / 0.7292                         & 31.32 / 0.8684                        \\
				
				TOFlow      & -                 & 27.98 / 0.7990                         & 33.08 / 0.9054                                                \\
				
				DUF          & 5.8 M         & 28.63 / 0.8251                         & -                                                                  \\
				RBPN  & 12.2 M          & 30.09 / 0.8590                         & 37.07 / 0.9435                                       \\
				EDVR-M     & 3.3 M                  & 30.53 / 0.8699                         & 37.09 / 0.9446                                            \\
				EDVR        & 20.6 M           & 31.09 / 0.8800                         & {37.61} / {0.9489}      \\
				PFNL   & 3.0 M               & 29.63 / 0.8502                         & 36.14 / 0.9363                        \\
				MuCAN        & -                     & 30.88 / 0.8750                         & 37.32 / 0.9465                                               \\
				BasicVSR & 6.3 M             & {31.42} / {0.8909}               & 37.18 / 0.9450                         \\
				IconVSR  & 8.7 M            & {\underline{31.67}} / {\underline{0.8948}}               & {37.47} / {0.9476}           
				\\
				VSR-Transformer  & 32.6 M            & {31.19} / {0.8815}               & \bf{37.71} / \bf{0.9494} 
				\\ \hline
				\textbf{\mbox{S2SVR} (Ours)} & 13.4 M            & \bf{31.96} / \bf{0.8988}               & \underline{37.63} / \underline{0.9490}                               
				\\\hline 
			\end{tabular}}
		\vspace{-2mm}
		\caption{Quantitative comparison (PSNR/SSIM) on the video SR dataset REDS4 and Vimeo-90K-T. Bold and underlined text indicate the best and the second-best performance, respectively.}
		\label{vsr_quan}
	\end{center}
	\vspace{-7.5mm}
\end{table}

%% file: tables/video_deblur.tex
\begin{table}[t]
\vspace{-1mm}
\resizebox{0.45\textwidth}{!}{
\centering
\begin{tabular}{l|r|c|c}
    \hline
    Methods      &Params   & PSNR (dB) & SSIM 
    \\
    \hline   
    Tao \textit{et al.} & -  &30.29 & 0.9014 \\
    Su \textit{et al.}   &15.30 M   &27.31 & 0.8255   \\
    Kim \textit{et al.}  & - &26.82 & 0.8245 \\
    Nah \textit{et al.}  & -  &29.97 & 0.8947  \\
    EDVR  & 23.6 M  &26.83 & 0.8426  \\
    STFAN  & 5.37 M   &28.59 & 0.8608\\
    TSP & 16.19 M &\underline{31.67} & \bf{0.9279} \\
    UHDVD & -  &31.33 & 0.9210 \\
    \hline
    \bf{S2SVR (Ours)} & 8.44 M  &\bf{31.81} & \underline{0.9231} \\
    \hline  
  \end{tabular}
}
\vspace{-2mm}
\caption{Video deblurring performance comparison and model parameter analysis on the GOPRO dataset~\cite{r74}.}
\vspace{-4mm}
\label{tab:GOPRO_result}
\end{table}

%% file: tables/vqe.tex
\begin{table*}[t]
	\renewcommand\arraystretch{1.3}
	\centering
	\Huge
	\resizebox{0.99\textwidth}{!}{
	\begin{threeparttable}
		
		\begin{tabular}{|c|c|l|c c| c c| c c| c c| c c|c c|c c|}
			
			\hline
			\multirow{2}{*}{QP} & \multicolumn{2}{c|}{\multirow{2}{*}{Approach}} & \multicolumn{2}{c|}{AR-CNN} & \multicolumn{2}{c|}{DnCNN} &  \multicolumn{2}{c|}{DS-CNN} & \multicolumn{2}{c|}{MFQE 1.0} &
			\multicolumn{2}{c|}{MFQE 2.0} &
			\multicolumn{2}{c|}{STDF-R3L} &
			\multicolumn{2}{c|}{{\textbf{S2SVR}}} \\ [-0.3em]
			
			& \multicolumn{2}{c|}{} & \multicolumn{2}{c|}{\cite{dong2015compression}}  & \multicolumn{2}{c|}{\cite{Zhang2017Beyond}} &  \multicolumn{2}{c|}{\cite{yang2018enhancing}} & 
			\multicolumn{2}{c|}{\cite{r13}} &
			\multicolumn{2}{c|}{\cite{r12}} &
			\multicolumn{2}{c|}{\cite{deng2020spatio}} &
			\multicolumn{2}{c|}{(\textbf{Ours})} \\
			
			\hline 
			\multirow{20}{*}{37} & \multicolumn{2}{c|}{Metrics} & \multicolumn{14}{c|}{\Huge $\Delta$PSNR / \Huge $\Delta$SSIM}  \\
			
			\cline{2-17}
			& \multirow{2}{*}{A} & \textit{Traffic}
			& \multicolumn{2}{c|}{0.239 / 47} & \multicolumn{2}{c|}{0.238 / 57}  & \multicolumn{2}{c|}{0.286 / 60} & \multicolumn{2}{c|}{0.497 / 90} & \multicolumn{2}{c|}{0.585 / 102} & \multicolumn{2}{c|}{0.730 / 115} & \multicolumn{2}{c|}{\textbf{0.851} / \textbf{138}}\\
			
			& & \textit{PeopleOnStreet}
			& \multicolumn{2}{c|}{0.346 / 75} & \multicolumn{2}{c|}{0.414 / 82}  & \multicolumn{2}{c|}{0.416 / 85} & \multicolumn{2}{c|}{0.802 / 137} & \multicolumn{2}{c|}{0.920 / 157} & \multicolumn{2}{c|}{1.250 / 196} & \multicolumn{2}{c|}{\textbf{1.385} / \textbf{216}} \\
			
			\cline{2-17}
			& \multirow{5}{*}{B} & \textit{Kimono}
			& \multicolumn{2}{c|}{0.219 / 65} & \multicolumn{2}{c|}{0.244 / 75}  & \multicolumn{2}{c|}{0.249 / 75} & \multicolumn{2}{c|}{0.495 / 113} & \multicolumn{2}{c|}{0.550 / 118} & \multicolumn{2}{c|}{0.850 / 161} & \multicolumn{2}{c|}{\textbf{1.055} / \textbf{195}} \\
			
			& & \textit{ParkScene}
			& \multicolumn{2}{c|}{0.136 / 38} & \multicolumn{2}{c|}{0.141 / 50}  & \multicolumn{2}{c|}{0.153 / 50} & \multicolumn{2}{c|}{0.391 / 103} & \multicolumn{2}{c|}{0.457 / 123} & \multicolumn{2}{c|}{0.590 / 147} & \multicolumn{2}{c|}{\textbf{0.649} / \textbf{165}}\\
			
			& & \textit{Cactus}
			& \multicolumn{2}{c|}{0.190 / 38} & \multicolumn{2}{c|}{0.195 / 48}  & \multicolumn{2}{c|}{0.239 / 58} & \multicolumn{2}{c|}{0.439 / 88} & \multicolumn{2}{c|}{0.501 / 100} & \multicolumn{2}{c|}{0.770 / 138} & \multicolumn{2}{c|}{\textbf{0.828} / \textbf{152}}\\
			
			& & \textit{BQTerrace}
			& \multicolumn{2}{c|}{0.195 / 28} & \multicolumn{2}{c|}{0.201 / 38}  & \multicolumn{2}{c|}{0.257 / 48} & \multicolumn{2}{c|}{0.270 / 48} & \multicolumn{2}{c|}{0.403 / 67} & \multicolumn{2}{c|}{0.630 / 106} & \multicolumn{2}{c|}{\textbf{0.654} / \textbf{115}}\\
			
			& & \textit{BasketballDrive}
			& \multicolumn{2}{c|}{0.229 / 55} & \multicolumn{2}{c|}{0.251 / 58}  & \multicolumn{2}{c|}{0.282 / 65} & \multicolumn{2}{c|}{0.406 / 80} & \multicolumn{2}{c|}{0.465 / 83} & \multicolumn{2}{c|}{0.750 / 123} & \multicolumn{2}{c|}{\textbf{0.972} / \textbf{157}}\\
			
			\cline{2-17}
			& \multirow{4}{*}{C} & \textit{RaceHorses}
			& \multicolumn{2}{c|}{0.219 / 43} & \multicolumn{2}{c|}{0.253 / 65}  & \multicolumn{2}{c|}{0.267 / 63} & \multicolumn{2}{c|}{0.340 / 55} & \multicolumn{2}{c|}{0.394 / 80} & \multicolumn{2}{c|}{0.550 / 135} & \multicolumn{2}{c|}{\textbf{0.854} / \textbf{203}}\\
			
			& & \textit{BQMall}
			& \multicolumn{2}{c|}{0.275 / 68} & \multicolumn{2}{c|}{0.281 / 68}  & \multicolumn{2}{c|}{0.330 / 80} & \multicolumn{2}{c|}{0.507 / 103} & \multicolumn{2}{c|}{0.618 / 120} & \multicolumn{2}{c|}{0.990 / 180} & \multicolumn{2}{c|}{\textbf{1.080} / \textbf{205}}\\
			
			& & \textit{PartyScene}
			& \multicolumn{2}{c|}{0.107 / 38} & \multicolumn{2}{c|}{0.131 / 48}  & \multicolumn{2}{c|}{0.174 / 58} & \multicolumn{2}{c|}{0.217 / 73} & \multicolumn{2}{c|}{0.363 / 118} &\multicolumn{2}{c|}{\textbf{0.680} / 194} & \multicolumn{2}{c|}{0.628 / \textbf{236}}\\
			
			& & \textit{BasketballDrill}
			& \multicolumn{2}{c|}{0.247 / 58} & \multicolumn{2}{c|}{0.331 / 68}  & \multicolumn{2}{c|}{0.352 / 68} & \multicolumn{2}{c|}{0.477 / 90} &\multicolumn{2}{c|}{0.579 / 120} &\multicolumn{2}{c|}{0.790 / 149} & \multicolumn{2}{c|}{\textbf{0.949} / \textbf{179}}\\
			
			\cline{2-17}
			& \multirow{4}{*}{D} & \textit{RaceHorses}
			& \multicolumn{2}{c|}{0.268 / 55} & \multicolumn{2}{c|}{0.311 / 73}  & \multicolumn{2}{c|}{0.318 / 75} & \multicolumn{2}{c|}{0.507 / 113} & \multicolumn{2}{c|}{0.594 / 143} & \multicolumn{2}{c|}{0.830 / 208} & \multicolumn{2}{c|}{\textbf{1.010} / \textbf{237}}\\
			
			& & \textit{BQSquare}
			& \multicolumn{2}{c|}{0.080 / 8} & \multicolumn{2}{c|}{0.129 / 18}  & \multicolumn{2}{c|}{0.201 / 38} & \multicolumn{2}{c|}{-0.010 / 15} & \multicolumn{2}{c|}{0.337 / 65} & \multicolumn{2}{c|}{0.640 / 125} & \multicolumn{2}{c|}{\textbf{0.886} / \textbf{141}}\\
			
			& & \textit{BlowingBubbles}
			& \multicolumn{2}{c|}{0.164 / 35} & \multicolumn{2}{c|}{0.184 / 58}  & \multicolumn{2}{c|}{0.228 / 68} & \multicolumn{2}{c|}{0.386 / 120} & \multicolumn{2}{c|}{0.533 / 170} & \multicolumn{2}{c|}{\textbf{0.740} / 226} & \multicolumn{2}{c|}{0.710 / \textbf{230}}\\
			
			& & \textit{BasketballPass}
			& \multicolumn{2}{c|}{0.259 / 58} & \multicolumn{2}{c|}{0.307 / 75}  & \multicolumn{2}{c|}{0.335 / 78} & \multicolumn{2}{c|}{0.628 / 138} & \multicolumn{2}{c|}{0.728 / 155} & \multicolumn{2}{c|}{1.080 / 212} & \multicolumn{2}{c|}{\textbf{1.110} / \textbf{222}}\\
			
			\cline{2-17}
			& \multirow{3}{*}{E} & \textit{FourPeople}
			& \multicolumn{2}{c|}{0.373 / 50} & \multicolumn{2}{c|}{0.388 / 60}  & \multicolumn{2}{c|}{0.459 / 70} & \multicolumn{2}{c|}{0.664 / 85} & \multicolumn{2}{c|}{0.734 / 95} & \multicolumn{2}{c|}{0.940 / 117} & \multicolumn{2}{c|}{\textbf{1.021} / \textbf{136}}\\
			
			& & \textit{Johnny}
			& \multicolumn{2}{c|}{0.247 / 10} & \multicolumn{2}{c|}{0.315 / 40}  & \multicolumn{2}{c|}{0.378 / 40} & \multicolumn{2}{c|}{0.548 / 55} & \multicolumn{2}{c|}{0.604 / 68} & \multicolumn{2}{c|}{0.810 / 88} & \multicolumn{2}{c|}{\textbf{0.976} / \textbf{120}}\\
			
			& & \textit{KristenAndSara}
			& \multicolumn{2}{c|}{0.409 / 50} & \multicolumn{2}{c|}{0.421 / 60}  & \multicolumn{2}{c|}{0.481 / 60} & \multicolumn{2}{c|}{0.655 / 75} & \multicolumn{2}{c|}{0.754 / 85} & \multicolumn{2}{c|}{0.970 / 96} & \multicolumn{2}{c|}{\textbf{1.035} / \textbf{113}}\\
			
			\cline{2-17}
			& \multicolumn{2}{c|}{Average}
			& \multicolumn{2}{c|}{0.233 / 45} & \multicolumn{2}{c|}{0.263 / 58}  & \multicolumn{2}{c|}{0.300 / 63} & \multicolumn{2}{c|}{0.455 / 88} & \multicolumn{2}{c|}{0.562 / 109} & \multicolumn{2}{c|}{0.830 / 151} & \multicolumn{2}{c|}{\textbf{0.925} / \textbf{176}}\\
			
			\hline 
			
		\end{tabular}
		
	\end{threeparttable}
	}
\vspace{-4mm}
\caption{Overall comparison of compressed video enhancement for $\Delta$PSNR (dB) and $\Delta$SSIM ($\times10^{-4}$) over test sequences at QP=37. We experiment with five different video resolutions: A (2,560×1,600), B (1,920×1,080), C (832×480), D (480×240), E (1,280×720).}
\label{tab:vqe}
\end{table*}

%% file: tables/ablation.tex
\begin{table}[t]
\footnotesize
\center
\begin{center}
\begin{tabular}{l|cc|cc}
\hline  
 \multirow{2}{*}{Method} &  \multicolumn{2}{c|}{RAFT} &  \multicolumn{2}{c}{Pwclite}
\\
\cline{2-5}
 & \multicolumn{1}{c|}{Sup.} & Unsup.&  \multicolumn{1}{c|}{Sup.} & Unsup.
\\
\hline 
Params & 4.81 M  & - & 2.24 M  & 2.24 M  
\\
\hline
PSNR (dB) & 31.88 & - & 31.79 & \textbf{31.96}
\\
\hline  
\end{tabular}
\end{center}
\vspace{-3mm}
\caption{Model analysis with supervised (Sup.) and unsupervised (Unsup.) optical flow estimation model training.}
\label{tab:unsupervised_optical}
\vspace{-3mm}
\end{table}

%% file: tables/Length.tex
\begin{table}[t]
\scriptsize
\footnotesize
\centering
\begin{center}
\begin{tabular}{c|c|c|c}
\hline
\multirow{1}{*}{Length} &  EDVR-M  &  EDVR  &  \textbf{S2SVR (Ours)}  
\\
\hline 
5 & 27.78  & 28.05  & 28.10 (+ 0.05)  
\\
15 & 28.47  & 28.80 & 29.25 (+ 0.45)  
\\

30 & 27.76 & 28.01 & 28.53 (+ 0.52)  \\
50 &  27.52 &  27.77 & 28.34 (\bf + 0.57) 
\\ 
\hline 
\end{tabular}
\end{center}
\vspace{-4.5mm}
\caption{Quantitative comparison on sequences of different length.}
\label{tab:length}
\vspace{-6mm}
\end{table}